\documentclass{article}
\usepackage{spconf,amsmath,graphicx,hyperref}
\usepackage{booktabs}
\usepackage{graphicx}
\usepackage{makecell}
\usepackage{tabularx}
\usepackage{hyperref}
\usepackage{xurl}
\usepackage{xcolor}

\newcolumntype{C}{>{\centering\arraybackslash}X}
\usepackage{float}

\title{TIDE: Trajectory-Informed Skill Distillation for Efficient Text-to-Video Prompt Optimization}
\name{Chengxing Zhou$^{1}$, \qquad Yuliang Huang$^{1}$, \qquad Jingwei Song$^{2,\dagger}$\thanks{$^{\dagger}$Corresponding author. Email: u3638265@connect.hku.hk}}
  
\address{$^{1}$ Sun Yat-sen University, China \qquad  $^{2}$The University of Hong Kong, China}

\begin{document}
\ninept
\maketitle
\begin{abstract}
Text-to-video (T2V) models are highly sensitive to prompt specificity, while iterative prompt optimization typically improves generation quality only through repeated video generation and vision-language model evaluation at inference time. We propose TIDE, a framework that distills this costly search process into an interpretable and reusable prompt-rewriting Skill. TIDE formulates prompt rewriting as a set of source-conditioned rule modules, each targeting a distinct semantic dimension. Once distilled, the frozen Skill activates the relevant rules and generates both positive and negative prompts in a single LLM call, followed by only one T2V generation, without test-time visual search or backbone fine-tuning. Evolved using only 100 prompts over 14 hours, TIDE achieves prompt-optimization performance comparable to computationally expensive search-based methods. Moreover, when transferred to previously unseen video-generation models, TIDE outperforms existing approaches specifically designed or trained for those models. Code is available \href{https://github.com/xiningin/TIDE_T2V}{here}.

\end{abstract}
\section{Introduction}
\label{sec:intro}

\label{sec:intro}

Video generation has advanced rapidly in recent years, catalyzed by the emergence of Sora~\cite{openai2024sora}. Building on this momentum, models~\cite{bytedance2026seedance25,minimax2026h3,googledeepmind2026veo,blackforestlabs2026flux3video} can generate visually compelling and temporally coherent videos from text. Their broad applicability—from advertising and creative design to film production and world simulation—has begun to yield real-world deployments and emerging commercial applications, underscoring the field's substantial practical potential. 


However, a gap remains between users' concise prompts and the detailed descriptions needed to fully exploit these models, motivating prompt optimization for video generation. One-shot methods use an LLM with a fixed prompting strategy or handcrafted templates to rewrite or augment prompts efficiently at inference time~\cite{ma2026pos,wang2025promptenhancer}. A complementary line amortizes optimization by training prompt rewriters offline on curated or optimized prompts using supervised, reinforcement-learning, or preference-based objectives~\cite{hao2023optimizing,cao2023beautifulprompt,ji2025prompt,cheng2025vpo,gao2025rapo}. These paradigms lack direct visual feedback at inference time and therefore cannot adapt to prompt-specific generation failures. More recent approaches perform dynamic, test-time optimization using feedback from generated visual content~\cite{manas2024improving,long2026vista}: they repeatedly generate candidates, assess their quality, and revise prompts until satisfactory results are obtained. In T2V generation, this feedback-driven process adapts to each prompt, is generally agnostic to the underlying generator, and often achieves stronger optimization than one-shot rewriting. These advantages come at a steep computational cost, however, as repeated video generation and model calls can take hours for a single prompt while producing only seconds of video.

This trade-off motivates our central question: \emph{Can the experience accumulated during iterative optimization be retained in a compact artifact, so that a subsequent one-shot rewrite can achieve the benefits of iterative search?} Rather than learning solely from offline optimized prompt pairs, we distill experience into an interpretable and reusable prompt-rewriting Skill. Unlike prompt-pair fine-tuning, which typically supervises only the initial and optimized endpoints, our approach also distills intermediate critiques, suggestions, candidate comparisons, and dimension-specific corrections from each trajectory, yielding a richer supervision signal and making fuller use of the collected optimization experience. The resulting Skill is model-agnostic and exposes its learned behavior as explicit rewriting rules, while requiring only one LLM call and one video generation at inference time. We call this framework \textbf{TIDE}, which organizes the acquired knowledge into distinct semantic dimensions, including motion specification, camera framing, identity preservation, and other semantic aspects. Accordingly, we formulate the rewriting policy as independent, source-triggered rule modules, each targeting a specific aspect of prompt quality. As illustrated in Fig.~\ref{fig:overview}, we continuously mine dimension-specific patches from online-generated optimization trajectories to update its rule modules and synchronize them with the Skill.

In summary, we introduce TIDE, a framework that distills iterative prompt-optimization experience into an interpretable, reusable, and inference-efficient Skill through modular, source-triggered rule updates. Distilled from optimization trajectories of only 100 prompts on a single T2V backbone, TIDE matches the source iterative optimizer and transfers effectively to other backbones, consistently outperforming prompt-optimization methods designed or fine-tuned for them.


\definecolor{module_1}{HTML}{6D8EBF}
\definecolor{module_2}{HTML}{D6B656}
\definecolor{module_3}{HTML}{559B67}
\begin{figure*}[t]
\centering
\includegraphics[
  width=\textwidth,
  trim=0.7cm 0.3cm 0.7cm 0.2cm,
  clip
]{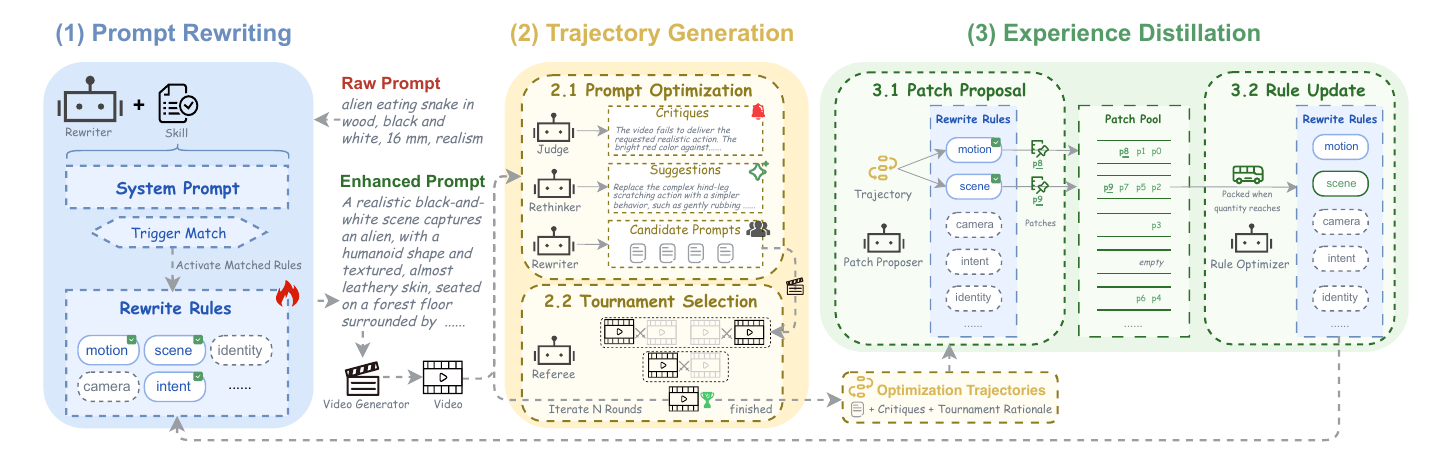}
\caption{Pipeline of the proposed \textbf{TIDE} framework. \textbf{\textcolor{module_1}{(1) Prompt Rewriting}}: Given a raw prompt—typically a short user query—the rewriter activates relevant Skill rules to produce an enhanced prompt. \textbf{\textcolor{module_2}{(2) Trajectory Generation}}: The generated video is critiqued, suggestions guide revised prompt candidates, and tournament selection chooses the next-round incumbent; this loop repeats to iteratively improve generation quality. \textbf{\textcolor{module_3}{(3) Experience Distillation}}: Prompt-optimization trajectories yield reusable, dimension-specific patches for the corresponding rule pools. Once full, a pool's patches update the associated rule and synchronize it with the Skill for subsequent rewriting.}
\label{fig:overview}
\end{figure*}

\section{Method}
\label{sec:method} 

\subsection{Skill Architecture}
Our central design premise is that the knowledge discovered during iterative test-time optimization decomposes into multiple, distinct semantic dimensions rather than forming a single monolithic rewriting policy. Concretely, the Skill comprises three conceptual components: (i) a system prompt, (ii) source-trigger specifications, and (iii) a collection of rule modules, as depicted in \textcolor{module_1}{stage (1)} of Fig.~\ref{fig:overview}. We encode these components at revision $t$ as the tuple
\begin{equation}
\mathcal{S}^{(t)}
=
\left(s_{\mathrm{sys}},
\{r_d^{(t)}\}_{d\in\mathcal{D}}\right),
r_d^{(t)}
=
\bigl(\tau_d,a_d,\iota_d,\nu_d,\kappa_d\bigr),
\label{eq:skill}
\end{equation}
where $s_{\mathrm{sys}}$ denotes the system prompt of the Skill, $\mathcal{D}$ denotes the set of semantic dimensions and $r_d^{(t)}$ is the Rule associated with dimension $d$. The trigger specifications $(\tau_d,a_d)$ implement Trigger Match, while $(\iota_d,\nu_d,\kappa_d)$ define the corresponding rewriting rule. More specifically, $\tau_d$ denotes a source-dependent trigger predicate, $a_d$ its responsibility axis, $\iota_d$ the rewriting instruction, $\nu_d$ a set of conditional artifact terms for negative prompting, and $\kappa_d$ the module's capability scope. The system prompt defines the global rewriting policy and output format, enforcing source fidelity and executable temporal--visual descriptions while keeping the output focused on visual content. This explicit design keeps the routing and rewriting processes interpretable and auditable.

Given a raw prompt $x$, Trigger Match evaluates each source-dependent trigger and activates the rules whose semantic dimensions are supported by $x$, together with the global rules. Let  $\mathcal{G}\subseteq\mathcal{D}$ denote the dimensions with globally active rules and let $\tau_d(x)\in\{0,1\}$ indicate whether the source satisfies the trigger of dimension $d$; the activated set is
\begin{equation}
\mathcal{A}(x)=\mathcal{G}\cup
\left\{d\notin\mathcal{G}:\tau_d(x)=1\right\}.
\label{eq:activation}
\end{equation}
Each conditional trigger is evaluated only along its designated axis, and multiple rules may be activated when the source supports multiple dimensions. The rewriter then conditions on the system prompt and the activated rules to produce an enhanced positive prompt and a rule-conditioned negative prompt:
\begin{equation}
\bigl(p^{+},\widetilde{n}\bigr)=
R_{\theta}^{\mathrm{LLM}}\!\left(x,s_{\mathrm{sys}},
\left\{r_d^{(t)}:d\in\mathcal{A}(x)\right\}\right).
\label{eq:rewrite}
\end{equation}
$p^{+}$ preserves all explicit source constraints and adds only
executable details supported by the activated instructions. Each activated rule
contributes its instruction $\iota_d$ to the rewrite and its artifact terms
$\nu_d$ to the rule-conditioned negative prompt. Following prior work on automated negative prompting~\cite{park2026guiding}, we use concise and non-redundant artifact phrases. The rule-conditioned negative prompt $\widetilde{n}$ is restricted to the allowed term set $\mathcal{N}(x)=\bigcup_{d\in\mathcal{A}(x)}\nu_d$ and capped at eight terms per rewrite. The LLM rewriter parameters remain frozen during Skill evolution; only the explicit Rule contents are updated from distilled optimization experience.

\subsection{Iterative Prompt-Optimization Trajectories}


Our trajectory-generation module, corresponding to \textcolor{module_2}{stage (2)} in Fig.~\ref{fig:overview}, adapts the prompt self-improvement procedure
of VISTA~\cite{long2026vista}, retaining its visual feedback loop while
discarding audio-specific components. Starting from the enhanced prompt
$p_i^{(0)}$ produced by the Skill, the text-to-video generator renders a video
at each optimization round. A group of judges critique the video's quality and
alignment with the source prompt. A re-thinker converts these critiques into
actionable suggestions, which guide the construction of multiple revised
prompt candidates. Their generated videos are compared in a pairwise
tournament, and the winning candidate becomes the incumbent for the next
round.

For every training prompt, we retain the complete optimization record as a
prompt-optimization trajectory:
\begin{equation}
\mathcal{T}_i=\left\{x_i,\left(p_i^{(k)},c_i^{(k)},s_i^{(k)},
\mathcal{P}_i^{(k)},w_i^{(k)}\right)_{k=1}^{K}\right\},
\label{eq:trajectory}
\end{equation}
where $x_i$ is the source prompt, $p_i^{(k)}$ the incumbent prompt,
$c_i^{(k)}$ the judges' critique, $s_i^{(k)}$ the re-thinker's suggestions,
$\mathcal{P}_i^{(k)}$ the revised candidate prompts, and $w_i^{(k)}$ the
tournament outcomes together with their decision rationales. These
trajectories are subsequently used to distill reusable rule patches that
encode dimension-specific optimization experience.

\subsection{Experience-Aggregated Rule Refinement}

Our experience-distillation module, corresponding to \textcolor{module_3}{stage (3)} in Fig.~\ref{fig:overview}, extracts reusable rule updates from the collected trajectories.
For each trajectory $\mathcal{T}_i$, let $\mathcal{A}_i=\mathcal{A}(x_i)$ denote
the dimensions activated for its source prompt. A Patch Proposer examines the
evidence along every activated dimension $d\in\mathcal{A}_i$ and determines whether it
supports a reusable modification of the corresponding rule $r_d^{(t)}$. When
such evidence is present, it emits one or more dimension-specific patches,
potentially spanning multiple dimensions and multiple corrections within the
same dimension. Each patch is associated with a trajectory--dimension pair
and must encode a generalizable correction within the rule's responsibility axis $a_d$ and capability scope $\kappa_d$, rather than memorize a particular prompt,
entity, video, seed, or benchmark example. The proposed change may refine the
rewriting instruction $\iota_d$ and the conditional artifact terms $\nu_d$;
the trigger $\tau_d$ is modified only when the trajectory provides explicit
evidence of a routing error.

To prevent noisy evidence from a single example from causing unstable or
over-specialized rule changes, accepted patches are stored in independent,
dimension-specific Patch Pools $\{Q_d\}$. Patches from different dimensions
are never mixed. A rule becomes eligible for refinement only after its pool
contains a complete batch of $B$ patches; the rule optimizer then synthesizes
the earliest batch into a complete replacement:
\begin{equation}
r_d^{(t+1)}=U\!\left(r_d^{(t)},Q_d[:B]\right),|Q_d|\geq B .
\label{eq:rule_update}
\end{equation}
After a successful update, the consumed patches are removed from $Q_d$.
If validation still fails after schema repair, the batch update is aborted:
the Rule is not applied and the selected patches remain in the pool. When multiple rules are eligible, the scheduler prioritizes the least recently updated rule (with never-updated rules first), preventing starvation and balancing updates across dimensions.

\section{Experiments}
\label{sec:experiments}

\subsection{Experiment Setting}
\label{sec:exp_setup}




\textbf{Implementation Details.}
We evolve the prompt-rewriting Skill with SANA-Video-2B~\cite{chen2026sanavideo}, generating 5-second, 480p videos with 81 frames. DeepSeek-V4-Flash~\cite{liu2024deepseek} handles text-only stages, including rewriting, candidate generation, trajectory distillation, and rule updates, while Qwen3.6-35B-A3B~\cite{yang2025qwen3} evaluates videos sampled at 2 FPS for quality assessment, critique generation, and tournament selection. During the Trajectory Generation stage, we perform two optimization rounds, generating four candidate prompts in each round; a rule is updated after four dimension-specific patches accumulate. The Skill is evolved on 100 prompts collected from EvalCrafter~\cite{liu2024evalcrafter}. Training takes 14 hours on four NVIDIA A100 GPUs, including 12.4 hours for trajectory collection.

\textbf{Evaluation Benchmarks.}
We draw evaluation prompts from VBench2~\cite{zheng2025vbench} and T2V-CompBench~\cite{sun2025t2v}. The VBench2 subset contains 431 canonical prompts across nine dimensions, while T2V-CompBench contains 200 prompts each for object interactions and generative numeracy. Unless otherwise specified, each prompt is rendered with three random seeds. For the matched comparison in Table~\ref{tab:vista_matched}, all methods use an 80-prompt subset sampled from VBench2 because of the high cost of iterative search. Within each backbone, all methods use identical generation settings and seeds. SANA-Video-2B and CogVideoX1.5-5B~\cite{yang2025cogvideox} use $832\times480$ resolution, 81 frames at 16 FPS, 50 denoising steps, and guidance scale 6.0. Wan2.2-TI2V-5B~\cite{wan2025wan} uses the same spatial, temporal, and denoising settings with its default guidance scale and sampling shift.

Across all tables, we abbreviate the VBench2 dimensions as SC (subject consistency), MS (motion smoothness), DD (dynamic degree), AQ (aesthetic quality), IQ (imaging quality), OC (overall consistency), MO (multiple objects), HA (human action), and SR (spatial relationship), and the T2V-CompBench dimensions as GN (generative numeracy) and OI (object interactions).

\textbf{Comparison Methods.}
We compare against VISTA~\cite{long2026vista} on the matched VBench2 subset using SANA-Video-2B, VPO-5B~\cite{cheng2025vpo} on CogVideoX1.5-5B, and \emph{Custom}~\cite{video_prompting_skill}, a community-authored Skill tailored to Wan2.2, on Wan2.2-TI2V-5B. For VISTA, we follow its original protocol of four optimization rounds and use the final video selected by its search. We additionally include each backbone's official prompt-extension procedure as a baseline to measure improvements over native prompt extension. For a fair comparison, all Skill-based rewriting uses DeepSeek-V4-Flash. To evaluate transferability, TIDE is trained only once with SANA-Video-2B and remains frozen when applied to other backbones.


\subsection{Main Results}
\label{sec:results}

\begin{table}[t]
\centering
\caption{Comparison of TIDE with multi-round iterative prompt optimization across the nine VBench2 dimensions. The average scores of Raw, VISTA, $\mathrm{Ours}_{\mathrm{init}}$, and $\mathrm{Ours}_{\mathrm{distil}}$ are 75.72, 78.37, 77.91, and 78.60, respectively. $\mathrm{Ours}_{\mathrm{init}}$ denotes the Skill before trajectory distillation, while $\mathrm{Ours}_{\mathrm{distil}}$ denotes the final Skill distilled from VISTA's optimization trajectories. The best and second-best results are highlighted in bold and underlined, respectively.}
\vspace{3pt}
\label{tab:vista_matched}

\setlength{\tabcolsep}{2.5pt}
\renewcommand{\arraystretch}{1.12}

\begin{tabularx}{\columnwidth}{@{}p{0.13\columnwidth}*{9}{C}@{}}
\toprule
\textbf{Method}
& \textbf{SC}
& \textbf{MS}
& \textbf{DD}
& \textbf{AQ}
& \textbf{IQ}
& \textbf{OC}
& \textbf{MO}
& \textbf{HA}
& \textbf{SR} \\
\midrule

Raw
& 95.64
& 98.55
& \textbf{87.50}
& \underline{61.25}
& \underline{67.79}
& \textbf{28.97}
& 78.91
& 81.25
& 81.59 \\

VISTA
& \textbf{96.60}
& \textbf{98.70}
& 75.00
& \textbf{61.28}
& \textbf{70.67}
& 27.61
& \textbf{98.83}
& 93.75
& 82.89 \\

\mbox{Ours$_{\mathrm{init}}$}
& 95.35
& 98.59
& \underline{79.17}
& 60.73
& 65.78
& 28.22
& \underline{92.45}
& \underline{97.92}
& \underline{82.94} \\

\mbox{Ours$_{\mathrm{distil}}$}
& \underline{96.22}
& \underline{98.63}
& 77.08
& 61.07
& 66.43
& \underline{28.24}
& 91.28
& \textbf{100.0}
& \textbf{88.43} \\

\bottomrule
\end{tabularx}
\end{table}

\begin{table*}[!t]
\centering
\caption{Comparison with other prompt optimization methods across CogVideoX1.5 and Wan2.2.}
\vspace{3pt}
\label{tab:cross_model}

{
\small
\setlength{\tabcolsep}{1.5pt}
\renewcommand{\arraystretch}{1.08}

\begin{tabularx}{\textwidth}{@{}p{0.10\textwidth}*{12}{C}@{}}
\toprule
&
&
\multicolumn{9}{c}{VBench2}
&
\multicolumn{2}{c}{T2VCompBench}
\\
\cmidrule(lr){3-11}
\cmidrule(lr){12-13}

Method
& \textbf{Avg.}
& \textbf{SC}
& \textbf{MS}
& \textbf{DD}
& \textbf{AQ}
& \textbf{IQ}
& \textbf{OC}
& \textbf{MO}
& \textbf{HA}
& \textbf{SR}
& \textbf{GN}
& \textbf{OI}
\\
\midrule

\multicolumn{13}{@{}l}{\textit{CogVideoX1.5-5B}} \\

Raw
& 54.83
& \textbf{96.18}
& \textbf{99.07}
& 16.20
& 56.66
& \textbf{62.35}
& 24.45
& 49.95
& 72.00
& 49.83
& 33.95
& 42.50
\\

Baseline
& 60.73
& 94.31
& 98.94
& 33.33
& \textbf{60.24}
& 61.65
& \textbf{26.35}
& 58.82
& 83.67
& \textbf{58.23}
& 34.80
& 57.75
\\

VPO
& 61.58
& 95.11
& 98.66
& \textbf{34.26}
& 58.31
& 60.36
& 26.06
& 61.94
& \textbf{92.33}
& 50.20
& 39.12
& \textbf{61.00}
\\

Ours
& \textbf{61.87}
& 95.50
& 98.96
& \textbf{34.26}
& 56.39
& 59.56
& 25.62
& \textbf{68.67}
& 91.00
& 51.75
& \textbf{42.96}
& 55.90
\\

\midrule

\multicolumn{13}{@{}l}{\textit{Wan2.2-TI2V-5B}} \\

Raw
& 53.69
& \textbf{97.16}
& 98.98
& 8.33
& 58.68
& 66.66
& 23.34
& 43.19
& 68.33
& 57.57
& 25.31
& 43.04
\\

Baseline
& 63.64
& 96.47
& 99.10
& 24.54
& 66.70
& \textbf{67.21}
& 25.46
& 64.00
& 88.67
& 66.19
& 29.96
& 71.78
\\

Custom
& 61.89
& 95.59
& 98.89
& 21.30
& 60.95
& 66.11
& \textbf{27.28}
& 66.59
& 95.33
& 61.52
& 24.60
& 62.62
\\

Ours
& \textbf{67.29}
& 96.11
& \textbf{99.03}
& \textbf{24.07}
& \textbf{62.91}
& 65.40
& 26.97
& \textbf{81.02}
& \textbf{97.33}
& \textbf{73.61}
& \textbf{40.62}
& \textbf{73.14}
\\

\bottomrule
\end{tabularx}
}
\end{table*}

\begin{table*}[!t]
\centering
  \caption{Ablation results for rule selection, negative-prompt distillation, and anti-starvation scheduling on Wan2.2.}
\label{tab:wan_ablation}
\vspace{3pt}

{
\small
\setlength{\tabcolsep}{1.5pt}
\renewcommand{\arraystretch}{1.08}

\begin{tabularx}{\textwidth}{@{}p{0.15\textwidth}*{12}{C}@{}}
\toprule
&
&
\multicolumn{9}{c}{VBench2}
&
\multicolumn{2}{c}{T2VCompBench}
\\
\cmidrule(lr){3-11}
\cmidrule(lr){12-13}

Method
& \textbf{Avg.}
& \textbf{SC}
& \textbf{MS}
& \textbf{DD}
& \textbf{AQ}
& \textbf{IQ}
& \textbf{OC}
& \textbf{MO}
& \textbf{HA}
& \textbf{SR}
& \textbf{GN}
& \textbf{OI}
\\
\midrule

Random Selection
& 66.76
& 94.94
& 98.94
& 31.94
& 62.12
& 64.42
& 27.07
& 77.57
& 96.67
& 73.81
& 35.34
& 71.54
\\

w/o Rule Selection
& 66.81
& 94.92
& 98.65
& 30.09
& \textbf{63.31}
& 65.36
& \textbf{27.60}
& 72.26
& 97.00
& \textbf{78.54}
& 34.99
& 72.21
\\

w/o NP Distil
& 65.46
& 94.89
& \textbf{99.25}
& 21.76
& 61.06
& 61.75
& 27.18
& 79.60
& 93.33
& 74.59
& 35.10
& 71.55
\\

w/o Anti-Starvation
& 66.97
& 94.63
& 98.90
& \textbf{35.65}
& 62.23
& 64.08
& 26.96
& 78.23
& 96.00
& 75.65
& 34.18
& 70.12
\\

Ours
& \textbf{67.30}
& \textbf{96.11}
& 99.03
& 24.07
& 62.91
& \textbf{65.40}
& 26.97
& \textbf{81.02}
& \textbf{97.33}
& 73.61
& \textbf{40.62}
& \textbf{73.14}
\\

\bottomrule
\end{tabularx}
}
\end{table*}

\textbf{Effective Distillation of Exploration Experience.} As shown in Table~\ref{tab:vista_matched}, even before trajectory distillation, $\mathrm{Ours}_{\mathrm{init}}$ improves the average score over Raw prompts (77.91 vs.\ 75.72), demonstrating the effectiveness of the initial rewriting Skill. After distilling the exploration experience from VISTA's optimization trajectories, $\mathrm{Ours}_{\mathrm{distil}}$ further improves the average score to 78.60, matching and slightly surpassing VISTA's 78.37. This result demonstrates that TIDE captures and distills the experience accumulated through VISTA's multi-round iterative exploration into reusable rewriting rules, transferring its optimization benefits to a compact, single-pass Skill. While VISTA and $\mathrm{Ours}_{\mathrm{distil}}$ exhibit complementary strengths across different dimensions, the distilled Skill achieves the best overall performance without requiring test-time visual search.

\textbf{Cross-Model Generalization and Advantages.} As shown in Table 2, the Skill trained on SANA-Video generalizes to Wan2.2 and CogVideoX1.5 without additional training, achieving the best overall performance for both backbones. The improvements are most evident on controllability-oriented dimensions, including multiple-object understanding, human action, and spatial relationships, while preserving competitive subject fidelity and motion quality. Compared with VPO and the manually customized Skill, our method delivers more balanced gains across quality and semantic control, demonstrating that exploration experience distilled into the Skill transfers effectively across video-generation architectures.

\begin{figure}[t]
\centering
\includegraphics[width=0.49\textwidth]{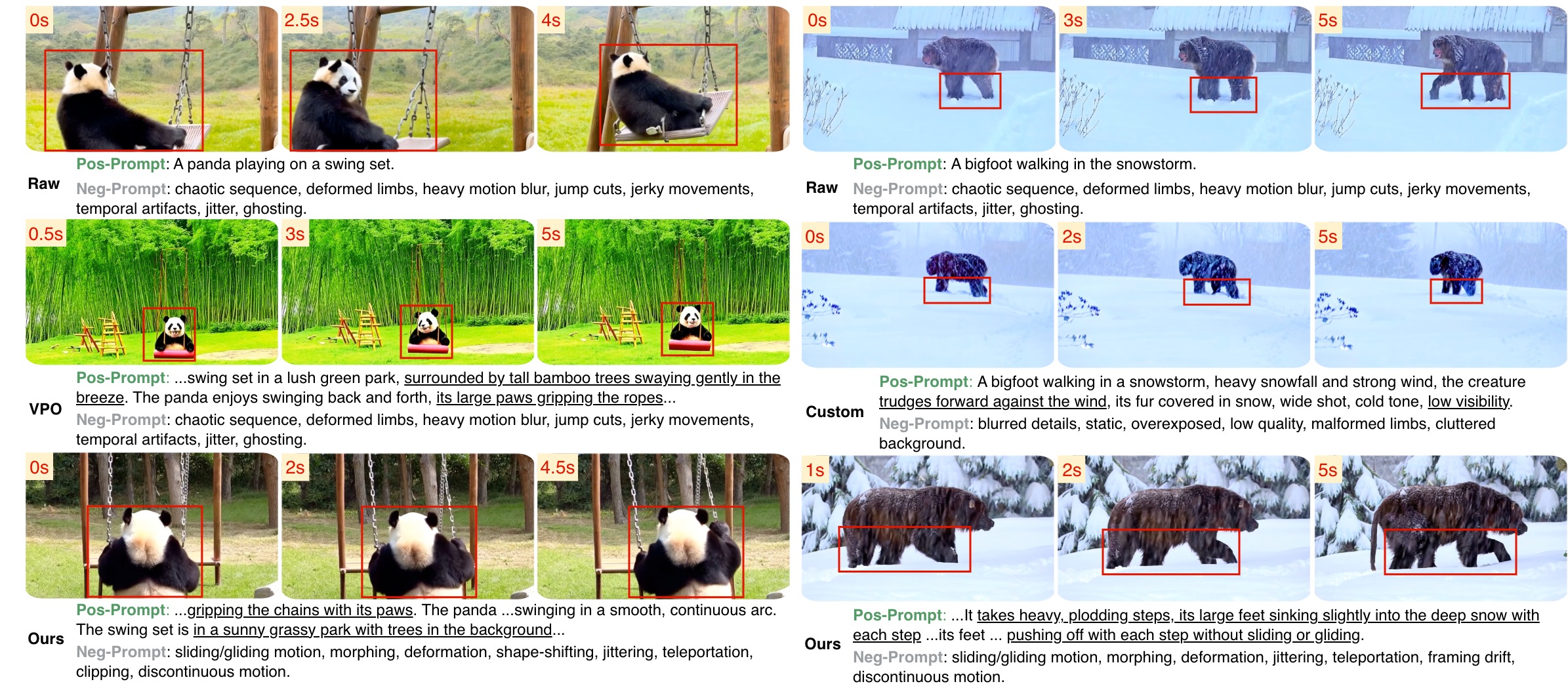}
\caption{Qualitative comparison of prompt optimization. Left: Ours and VPO on CogVideoX1.5. Right: Ours and Custom on Wan2.2.}
\label{fig:cases}
\end{figure}

\textbf{Qualitative Results.} Fig.~\ref{fig:cases} compares the optimized prompts and visual results. For ``A panda playing on a swing set,'' both VPO and Ours add a gripping action to make the panda--swing interaction more plausible. However, Ours better preserves the focus of the raw prompt, keeping the panda and swing visually dominant while achieving cleaner composition and higher visual quality. For ``A bigfoot walking in the snowstorm,'' the Raw result inadequately expresses the intended walking action, with noticeable leg movement appearing only near the end of the video. Both Custom and Ours address this issue by strengthening the motion description: Custom specifies that the creature ``trudges forward against the wind,'' while Ours further describes heavy steps and explicit foot--snow interaction, producing a more continuous and grounded walking motion. The targeted negative prompts suppress structural and temporal artifacts, improving subject consistency and motion smoothness.

\subsection{Ablation Studies}
\label{sec:ablation}
We ablate three key components of TIDE: rule selection, negative-prompt distillation, and anti-starvation scheduling. For rule selection, ``w/o Rule Selection'' activates all rules for every prompt, while ``Random Selection'' retains global rules and randomly samples the remaining rules under a budget of three. ``w/o NP Distil'' replaces dimension-conditioned negative prompts with a shared fixed one, and ``w/o Anti-Starvation'' removes recency-based prioritization during rule updates. All variants use the same settings as the full Skill.

As shown in Table~\ref{tab:wan_ablation}, Ours achieves the highest average score of 67.30, compared with 66.76 for Random Selection, 66.81 for w/o Rule Selection, 65.46 for w/o NP Distil, and 66.97 for w/o Anti-Starvation. The largest drop occurs without negative-prompt distillation (-1.84), with notable degradation in imaging quality, human action, generative numeracy, and object interaction. Although the rule-selection variants improve some dynamic-degree or spatial-relation scores, their lower averages indicate a weaker balance across dimensions. Similarly, removing anti-starvation increases dynamic degree but reduces the average by 0.33, confirming that balanced rule updates benefit overall performance.


\section{CONCLUSION}
\label{sec:conclusion}

We introduce TIDE, a trajectory-informed framework that distills iterative text-to-video prompt optimization into an interpretable, reusable rewriting skill. By distilling iterative visual search, TIDE alleviates this trade-off: it retains the strong performance and model-agnostic applicability of iterative methods while replacing costly multi-round generation, evaluation, and revision with a single LLM rewrite and T2V generation, matching search-based optimization from only 100 prompts on a single backbone and surpassing methods designed or fine-tuned for unseen generators. Despite these advantages, TIDE still has limitations. Distilled from trajectories collected on a single backbone with a small prompt set, its rules leave rarely triggered dimensions under-refined and may encode backbone- or judge-specific biases. Future work will therefore explore learning from more diverse optimization trajectories to improve robustness, and extend TIDE to audio-aware video optimization.

\vfill\pagebreak




\bibliographystyle{IEEEbib}
\bibliography{strings,refs}

\end{document}